\documentclass[11pt]{article}
\usepackage{acl2016}
\usepackage{times}

\usepackage{amsmath,amssymb,latexsym}
\usepackage{graphicx}
\usepackage{xcolor}

\usepackage[utf8]{inputenc}

\usepackage{enumitem}
\usepackage{multirow}
\usepackage{arydshln}
\usepackage[normalem]{ulem}

\makeatletter
\newcommand{\@BIBLABEL}{\@emptybiblabel}
\newcommand{\@emptybiblabel}[1]{}
\makeatother
\usepackage{hyperref}

\newcommand\Tstrut{\rule{0pt}{2.3ex}}       
\newcommand\Bstrut{\rule[-0.9ex]{0pt}{0pt}} 

\def\entr#1{\textcolor{purple}{\uline{#1}}}

\aclfinalcopy 

\title{A Context-aware Natural Language Generator for Dialogue Systems}

\author{Ondřej Dušek \and Filip Jurčíček \\
Charles University in Prague, Faculty of Mathematics and Physics \\
Institute of Formal and Applied Linguistics \\
Malostranské náměstí 25, CZ-11800 Prague, Czech Republic \\
\url{{odusek,jurcicek}@ufal.mff.cuni.cz}
}

\date{}

\begin{document}
\maketitle

\begin{abstract}
We present a novel natural language generation system for spoken dialogue systems 
capable of entraining (adapting) to users' way of speaking, providing contextually
appropriate responses.
The generator is based on recurrent neural networks and the sequence-to-sequence approach.
It is fully trainable from data which include preceding context along with responses to be generated.
We show that the context-aware generator yields significant improvements over the baseline in both
automatic metrics and a human pairwise preference test.
\end{abstract}

\section{Introduction}\label{sec:intro}

In a conversation, speakers are influenced by previous utterances of their counterparts and 
tend to adapt (align, entrain) their way of speaking to each other, 
reusing lexical items as well as syntactic structure \cite{reitter_computational_2006}.
Entrainment occurs naturally and subconsciously, facilitates successful conversations \cite{friedberg_lexical_2012,nenkova_high_2008}, and forms a natural source of variation in dialogues.
In spoken dialogue systems (SDS), users were reported to entrain to system prompts \cite{parent_lexical_2010}. 

\begin{figure}[tb]
\begin{center}
\includegraphics[width=7.8cm]{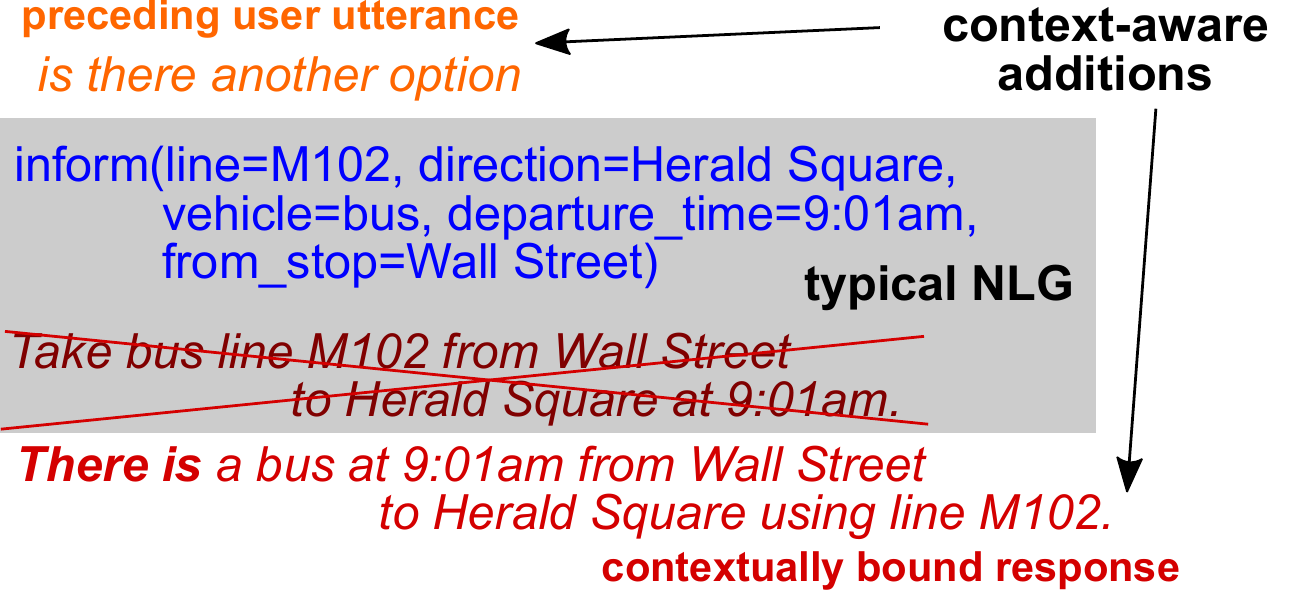}
\end{center}
\caption{An example of NLG input and output, with context-aware additions.}\label{fig:example}
\end{figure}

The function of natural language generation (NLG) components in task-oriented SDS 
typically is to produce a natural language sentence from a
\emph{dialogue act} (DA) \cite{young_hidden_2010} 
representing an action, such as \emph{inform} or \emph{request},
along with one or more attributes (\emph{slots}) and their values (see Fig.~\ref{fig:example}).
NLG is an important component of SDS which has a great impact on the perceived naturalness of the system; its quality can also influence the overall task success \cite{stoyanchev_lexical_2009,lopes_automated_2013}.
However, typical NLG systems in SDS only take the input DA into account and have no way of adapting to the user's way of speaking.
To avoid repetition and add variation into the outputs, they typically alternate between a handful of preset variants \cite{jurcicek_alex_2014} or use overgeneration and random sampling from a $k$-best list of outputs \cite{wen_semantically_2015}.
There have been several attempts at introducing entrainment into NLG in SDS, but they are limited to rule-based systems (see Section~\ref{sec:related}).

We present a novel, fully trainable context-aware NLG system for SDS that is able to entrain to the user and provides naturally variable outputs because generation is conditioned not only on the input DA, but also on the preceding user utterance (see Fig.~\ref{fig:example}).
Our system is an extension of \newcite{dusek_sequence--sequence_2016}'s generator based on sequence-to-sequence (seq2seq) models with attention \cite{bahdanau_neural_2014}.
It is, to our knowledge, the first fully trainable entrainment-enabled NLG system for SDS.
We also present our first results on the dataset of \newcite{dusek_context-aware_2016}, which includes the preceding user utterance along with each data instance (i.e., pair of input meaning representation and output sentence), and we show that our context-aware system outperforms the baseline in both automatic metrics and a human pairwise preference test.

In the following, we first present the architecture of our generator (see Section~\ref{sec:our-generator}), then give an account of our experiments in Section~\ref{sec:experiments}. We include a brief survey of related work in Section~\ref{sec:related}. Section~\ref{sec:concl} contains concluding remarks and plans for future work.

\section{Our generator}\label{sec:our-generator}

Our seq2seq generator is an improved version of \newcite{dusek_sequence--sequence_2016}'s generator, which itself is based on the
seq2seq model with attention \cite[see Fig.~\ref{fig:seq2seq}]{bahdanau_neural_2014} as implemented in the TensorFlow framework \cite{tensorflow2015-whitepaper}.\footnote{See \cite{dusek_sequence--sequence_2016} and \cite{bahdanau_neural_2014} for a more formal description of the base model.} We first describe the base model in Section~\ref{sec:seq2seq}, then list our context-aware improvements in Section~\ref{sec:context}.

\subsection{Baseline Seq2seq NLG with Attention}\label{sec:seq2seq}

\begin{figure*}[tb]
\begin{center}
\includegraphics[width=\textwidth]{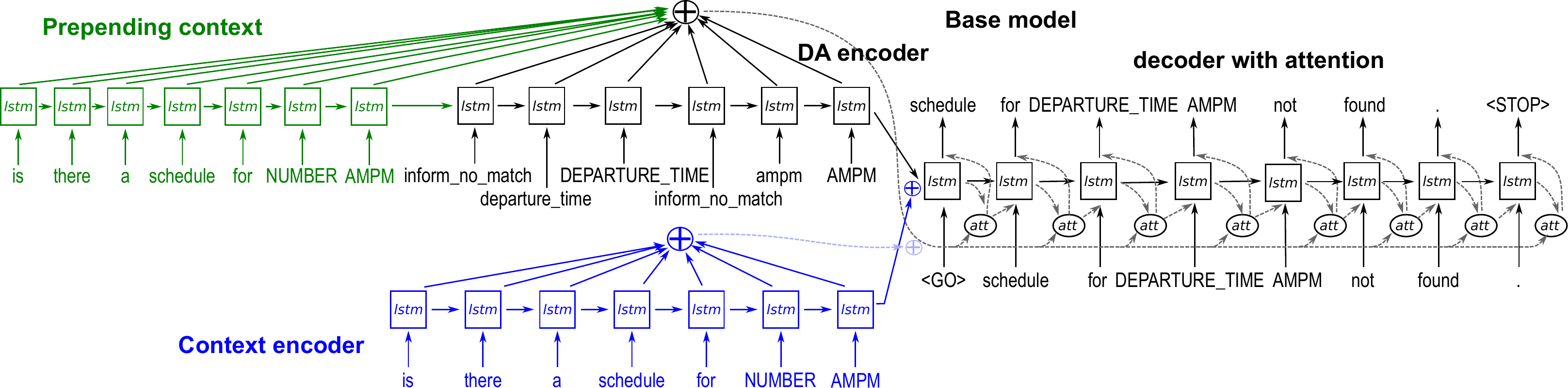}
\end{center}
\caption{The base Seq2seq generator (black) with our improvements: prepending context (green) and separate context encoder (blue).}\label{fig:seq2seq}
\vspace{-4mm}
\end{figure*}

The generation has two stages: The first, \emph{encoder stage} uses a recurrent neural network (RNN) composed of long-short-term memory (LSTM) cells \cite{hochreiter_long_1997,graves_generating_2013}
to encode a sequence of input tokens\footnote{Embeddings are used \cite{bengio_neural_2003}, i.e., $x_t$ and $y_t$ are vector representations of the input and output tokens.} $\mathbf{x}=\{x_1, \dots,x_n\}$ into a sequence of hidden states $\mathbf{h} =\{h_1,\dots,h_n\}$: 
\begin{equation}
h_t = \mathrm{lstm}(x_t, h_{t-1})
\end{equation}
The second, \emph{decoder stage} then uses the hidden states $\mathbf{h}$ to generate the output sequence $\mathbf{y} = \{y_1,\dots,y_m\}$. Its main component is a second LSTM-based RNN, which works over its own internal state $s_t$ and the previous output token $y_{t-1}$: 
\begin{equation}
s_t    = \mbox{lstm}((y_{t-1}\circ c_t)W_S,s_{t-1})\label{eq:s_t}
\end{equation}
It is initialized by the last hidden encoder state ($s_0 = h_n$) and a special starting symbol. The generated output token $y_t$ is selected from a softmax distribution:
\begin{equation}
p(y_t|y_{t-1}\dots,\mathbf{x}) = \mbox{softmax}((s_t\circ c_t)W_Y)\label{eq:y_t}
\end{equation}
In (\ref{eq:s_t}) and (\ref{eq:y_t}), $c_t$ represents the \emph{attention model} – a sum over all encoder hidden states, weighted by a feed-forward network with one $\tanh$ hidden layer; $W_S$ and $W_Y$ are linear projection matrices and “$\circ$” denotes concatenation.

DAs are represented as sequences on the encoder input: a triple of the structure “DA type, slot, value” is created for each slot in the DA and the triples are concatenated (see Fig.~\ref{fig:seq2seq}).
\footnote{While the sequence encoding may not necessarily be the best way to obtain a vector representation of DA, it was shown to work well \cite{dusek_sequence--sequence_2016}.}
The generator supports greedy decoding as well as beam search which keeps track of top $k$ most probable output sequences at each time step \cite{sutskever_sequence_2014,bahdanau_neural_2014}.

The generator further features a simple content classification reranker to penalize irrelevant or missing information on the output. It uses an LSTM-based RNN to encode the generator outputs token-by-token into a fixed-size vector. This is then fed to a sigmoid classification layer that outputs a 1-hot vector indicating the presence of all possible DA types, slots, and values. The vectors for all $k$-best generator outputs are then compared to the input DA and the number of missing and irrelevant elements is used to rerank them.

\subsection{Making the Generator Context-aware}\label{sec:context}

We implemented three different modifications to our generator that make its output dependent on the preceding context:
\footnote{For simplicity, we kept close to the basic seq2seq architecture of the generator; other possibilities for encoding the context, such as convolution and/or max-pooling, are possible.}

\paragraph{Prepending context.} The preceding user utterance is simply prepended to the DA and fed into the encoder (see Fig.~\ref{fig:seq2seq}). The dictionary for context utterances is distinct from the DA tokens dictionary.

\paragraph{Context encoder.} We add another, separate encoder for the context utterances. The hidden states of both encoders are concatenated, and the decoder then works with double-sized vectors both on the input and in the attention model (see Fig.~\ref{fig:seq2seq}).

\paragraph{$n$-gram match reranker.} We added a second reranker for the $k$-best outputs of the generator that promotes outputs that have a word or phrase overlap with the context utterance. We use geometric mean of modified $n$-gram precisions (with $n\in\{1,2\}$) as a measure of context overlap, i.e., BLEU-2 \cite{papineni_bleu:_2002} without brevity penalty. The log probability $l$ of an output sequence on the generator $k$-best list is updated as follows:
\begin{equation}
l = l + w\cdot\sqrt{p_1p_2} \label{eq:context} 
\end{equation}
In (\ref{eq:context}), $p_1$ and $p_2$ are modified unigram and bigram precisions of the output sequence against the context, and $w$ is a preset weight.
We believe that any reasonable measure of contextual match would be viable here, and we opted for modified $n$-gram precisions because of simple computation, well-defined range, and the relation to the de facto standard BLEU metric.\footnote{We do not use brevity penalty as we do not want to demote shorter output sequences. However, adding it to the formula in our preliminary experiments yielded similar results to the ones presented here.}
We only use unigrams and bigrams to promote especially the reuse of single words or short phrases.

In addition, we combine the $n$-gram match reranker with both of the two former approaches.

We used gold-standard transcriptions of the immediately preceding user utterance in our experiments in order to test the context-aware capabilities of our system in a stand-alone setting; in a live SDS, 1-best speech recognition hypotheses and longer user utterance history can be used with no modifications to the architecture.

\section{Experiments}\label{sec:experiments}

\begin{table}[tb]
\begin{tabular}{lcc}
\bf Setup                           & \bf BLEU & \bf NIST \\\hline
Baseline (context not used)         & 66.41    & 7.037 \Tstrut\\ 
$n$-gram match reranker             & 68.68    & 7.577 \\ 
Prepending context                  & 63.87    & 6.456 \\ 
\quad + $n$-gram match reranker     & 69.26    & 7.772 \\ 
Context encoder                     & 63.08    & 6.818 \\ 
\quad + $n$-gram match reranker     & 69.17    & 7.596 \\ 
\end{tabular}
\caption{BLEU and NIST scores of different generator setups on the test data.}\label{tab:results}
\vspace{-4mm}
\end{table}

We experiment on the publicly available dataset of \newcite{dusek_context-aware_2016}\footnote{The dataset is released at \url{http://hdl.handle.net/11234/1-1675}; we used a more recent version from GitHub (\href{https://github.com/UFAL-DSG/alex_context_nlg_dataset}{\texttt{https://github.com/UFAL-DSG/alex\_}}\linebreak\href{https://github.com/UFAL-DSG/alex_context_nlg_dataset}{\texttt{context\_nlg\_dataset}}), which contains several small fixes.} for NLG in the public transport information domain, which includes preceding context along with each pair of input DA and target natural language sentence.  It contains over 5,500 utterances, i.e., three paraphrases for each of the over 1,800 combinations of input DA and context user utterance. The data concern bus and subway connections on Manhattan, and comprise four DA types (\emph{iconfirm}, \emph{inform}, \emph{inform\_no\_match}, \emph{request}).
They are delexicalized for generation to avoid sparsity, i.e., stop names, vehicles, times, etc., are replaced by placeholders \cite{wen_stochastic_2015}.
We applied a 3:1:1 split of the set into training, development, and test data. We use the three paraphrases as separate instances in training data, but they serve as three references for a single generated output in validation and evaluation.

We test the three context-aware setups described in Section~\ref{sec:context} and their combinations, and we compare them against the baseline non-context-aware seq2seq generator.
Same as \newcite{dusek_sequence--sequence_2016}, we train the seq2seq models by minimizing cross-entropy on the training set using the Adam optimizer \cite{kingma_adam:_2015}, and we measure BLEU on the development set after each pass over the training data, selecting the best-performing parameters.\footnote{Based on our preliminary experiments on development data, we use embedding size 50, LSTM cell size 128, learning rate 0.0005, and batch size 20. Training is run for at least 50 and up to 1000 passes, with early stopping if the top 10 validation BLEU scores do not change for 100 passes.}
The content classification reranker is trained in a similar fashion, measuring misclassification on both training and development set after each pass.\footnote{We use the same settings except for the number of passes over the training data, which is at least 20 and 100 at most. For validation, development set is given 10 times more importance than the training set.}
We use 5 different random initializations of the networks and average the results.

Decoding is run with a beam size of 20 and the penalty weight for content classification reranker set to 100. We set the $n$-gram match reranker weight based on experiments on development data.\footnote{$w$ is set to 5 when the $n$-gram match reranker is run by itself or combined with the separate encoder, 10 if combined with prepending context.}

\subsection{Evaluation Using Automatic Metrics}

\begin{table*}[tb]\scriptsize
\begin{tabular}{ll}

Context & is there a later option \\
Input DA & iconfirm(alternative=next) \\\hdashline[0.5pt/2pt]
Baseline & Next connection. \\
$n$-gram match reranker & You want \entr{a later} connection. \\
Prepending context + $n$-gram match reranker & You want \entr{a later} connection. \\
Context encoder +  $n$-gram match reranker & You want \entr{a later option}. \Bstrut \\\hline

Context & i need to find a bus connection \Tstrut \\
Input DA & inform\_no\_match(vehicle=bus) \\\hdashline[0.5pt/2pt]
Baseline & No bus found, sorry. \\
$n$-gram match reranker & I did not \entr{find a bus} route. \\
Prepending context + $n$-gram match reranker & I'm sorry, I cannot \entr{find a bus connection}. \\
Context encoder +  $n$-gram match reranker & I'm sorry, I cannot \entr{find a bus connection}. \Bstrut \\\hline

Context & i rather take the bus \Tstrut \\
Input DA & inform(vehicle=bus, departure\_time=8:01am, direction=Cathedral Parkway, from\_stop=Bowling Green, \hbox to 0.37cm{line=M15)} \\\hdashline[0.5pt/2pt]
Baseline & At 8:01am by bus line M15 from Bowling Green to Cathedral Parkway. \\
$n$-gram match reranker & At 8:01am by bus line M15 from Bowling Green to Cathedral Parkway. \\
Prepending context + $n$-gram match reranker & You can \entr{take the M15 bus} from Bowling Green to Cathedral Parkway at 8:01am. \\
Context encoder +  $n$-gram match reranker & At 8:01am by bus line M15 from Bowling Green to Cathedral Parkway. \\

\end{tabular}
\caption{Example outputs of the different setups of our generator (with entrainment highlighted)}\label{tab:outputs}
\vspace{-4mm}
\end{table*}

Table~\ref{tab:results} lists our results on the test data in terms of the BLEU and NIST metrics \cite{papineni_bleu:_2002,doddington_automatic_2002}.
We can see that while the $n$-gram match reranker brings a BLEU score improvement, using context prepending or separate encoder results in scores lower than the baseline.\footnote{In our experiments on development data, all three methods brought a mild BLEU improvement.}
However, using the $n$-gram match reranker together with context prepending or separate encoder brings significant improvements of about 2.8 BLEU points in both cases, better than using the $n$-gram match reranker alone.\footnote{Statistical significance at 99\% level has been assessed using pairwise bootstrap resampling \cite{koehn_statistical_2004}.}
We believe that adding the context information into the decoder does increase the chances of contextually appropriate outputs appearing on the decoder $k$-best lists, but it also introduces a lot more uncertainty and therefore, the appropriate outputs may not end on top of the list based on decoder scores alone. The $n$-gram match reranker is then able to promote the relevant outputs to the top of the $k$-best list. However, if the generator itself does not have access to context information, the $n$-gram match reranker has a smaller effect as contextually appropriate outputs may not appear on the $k$-best lists at all.
A closer look at the generated outputs confirms that entrainment is present in sentences generated by the context-aware setups (see Fig.~\ref{tab:outputs}).

In addition to BLEU and NIST scores, we measured the slot error rate ERR \cite{wen_semantically_2015}, i.e., the proportion of missing or superfluous slot placeholders in the delexicalized generated outputs. For all our setups, ERR stayed around 3\%.

\subsection{Human Evaluation}

We evaluated the best-performing setting based on BLEU/NIST scores, i.e., prepending context with $n$-gram match reranker, in a blind pairwise preference test with untrained judges recruited on the CrowdFlower crowdsourcing platform.\footnote{\url{http://crowdflower.com}} 
The judges were given the context and the system output for the baseline and the context-aware system, and they were asked to pick the variant that sounds more natural. We used a random sample of 1,000 pairs of different system outputs over all 5 random initializations of the networks, and collected 3 judgments for each of them.
The judges preferred the context-aware system output in 52.5\% cases, significantly more than the baseline.\footnote{The result is statistically significant at 99\% level according to the pairwise bootstrap resampling test.}

We examined the judgments in more detail and found three probable causes for the rather small difference between the setups. First, both setups' outputs fit the context relatively well in many cases and the judges tend to prefer the overall more frequent variant (e.g., for the context “starting from Park Place”, the output “Where do you want to go?” is preferred over “Where are you going to?”).
Second, the context-aware setup often selects a shorter response that fits the context well (e.g., “Is there an option at 10:00 am?” is confirmed simply with “At 10:00 am.”), but the judges seem to prefer the more eloquent variant. And third, both setups occasionally produce non-fluent outputs, which introduces a certain amount of noise.

\section{Related Work}\label{sec:related}

Our system is an evolutionary improvement over the LSTM seq2seq system of \newcite{dusek_sequence--sequence_2016} and as such,
it is most related in terms of architecture to other
recent RNN-based approaches to NLG, which are not context-aware: RNN generation with a convolutional reranker by \newcite{wen_stochastic_2015}
and an improved LSTM-based version \cite{wen_semantically_2015}, as well as the LSTM encoder-aligner-decoder NLG system of \newcite{mei_what_2015}.
The recent end-to-end trainable SDS of \newcite{wen_network-based_2016} does have an implicit access to previous context, but the authors do not focus on its influence on the generated responses.

There have been several attempts at modelling entrainment in dialogue \cite{brockmann_modelling_2005,reitter_computational_2006,buschmeier_modelling_2010}
and even successful implementations of entrainment models in NLG systems 
for SDS, where entrainment caused an increase in perceived naturalness of the system responses \cite{hu_entrainment_2014} or increased naturalness and task success (Lopes et al., \nobrashortcite{lopes_automated_2013}; \nobracite{lopes_rule-based_2015}). 
However, all of the previous approaches are completely or partially rule-based. Most of them attempt to model entrainment
explicitly, focus on specific entrainment phenomena only, and/or require manually selected lists of variant expressions, while our system learns synonyms and entrainment rules implicitly from the corpus.
A direct comparison with previous entrainment-capable NLG systems for SDS is not possible in our stand-alone setting since their rules involve the history of the whole dialogue whereas we focus on the preceding utterance in our experiments.

\section{Conclusions and Further Work}\label{sec:concl}

We presented an improvement to our natural language generator based on the sequence-to-sequence approach \cite{dusek_sequence--sequence_2016}, allowing it to exploit preceding context user utterances to adapt (entrain) to the user's way of speaking and provide more contextually accurate and less repetitive responses.
We used two different ways of feeding previous context into the generator and a reranker based on $n$-gram match against the context.
Evaluation on our context-aware dataset \cite{dusek_context-aware_2016} showed a significant BLEU score improvement for the combination of the two approaches, which was confirmed in a subsequent human pairwise preference test.
Our generator is available on GitHub at the following URL:
\begin{center}
\url{https://github.com/UFAL-DSG/tgen}
\end{center}

In future work, we plan on improving the $n$-gram matching metric to allow fuzzy matching (e.g., capturing different forms of the same word), experimenting with more ways of incorporating context into the generator, controlling the output eloquence and fluency, and most importantly, evaluating our generator in a live dialogue system. We also intend to evaluate the generator with automatic speech recognition hypotheses as context and modify it to allow $n$-best hypotheses as contexts. Using our system in a live SDS will also allow a comparison against previous handcrafted entrainment-capable NLG systems.

\section*{Acknowledgments}

This work was funded by the Ministry of Education, Youth and Sports of the Czech Republic
under the grant agreement LK11221 and core research funding, SVV project 260~333, and GAUK
grant 2058214 of Charles University in Prague. It used language resources stored and distributed by
the LINDAT/CLARIN project of the Ministry of Education, Youth and Sports of the Czech Republic (project LM2015071).
The authors would like to thank Ondřej Plátek and Miroslav Vodolán for helpful comments.

\bibliographystyle{acl2016}
\bibliography{paper}

\end{document}